\documentclass{article}

\usepackage[preprint]{corl_2025} 
\usepackage{xcolor}

\usepackage[utf8]{inputenc} 
\usepackage[T1]{fontenc}    
\usepackage{hyperref}       
\usepackage{url}            
\usepackage{booktabs}       
\usepackage{amsfonts}       
\usepackage{nicefrac}       
\usepackage{microtype}      
\usepackage{graphicx}
\usepackage{hyperref}       
\usepackage{url}            
\usepackage{booktabs}       
\usepackage{amsfonts}       
\usepackage{nicefrac}       
\usepackage{microtype}      
\usepackage{bm}
\usepackage{amsmath}
\usepackage{amssymb}
\usepackage{mathtools}
\usepackage{amsthm}
\usepackage{multirow}
\usepackage{gensymb}

\usepackage[linesnumbered,ruled,vlined]{algorithm2e}
\usepackage[ruled]{algorithm2e}
\usepackage{algpseudocode}  
\usepackage{amsmath} 
\usepackage{caption}
\usepackage{subfigure}
\usepackage{graphicx}
\usepackage{svg}
\usepackage{pifont}
\usepackage{natbib}
\usepackage{authblk}
\title{Object-Focus Actor for Data-efficient Robot Generalization Dexterous Manipulation}

\renewcommand{\thefootnote}{}
\author[1$\dagger$]{Yihang Li}
\author[1$\dagger$]{Tianle Zhang}
\author[1]{Xuelong Wei}
\author[2]{Jiayi Li}
\author[1]{Lin Zhao}
\author[1]{Dongchi Huang}
\author[1]{Zhirui Fang}
\author[2]{Minhua Zheng}
\author[1]{Wenjun Dai}
\author[1]{Xiaodong He}

\affil[1]{JD Explore Academy, JD Company}
\affil[2]{Beijing Jiaotong University}

\begin{document}
\maketitle
\footnotetext{$\dagger$~These authors contributed equally to this work.}
\renewcommand{\thefootnote}{\arabic{footnote}}
\vspace{-0.2in}
\begin{abstract}

Robot manipulation learning from human demonstrations offers a rapid means to acquire skills but often lacks generalization across diverse scenes and object placements. This limitation hinders real-world applications, particularly in complex tasks requiring dexterous manipulation. Vision-Language-Action (VLA) paradigm leverages large-scale data to enhance generalization. However, due to data scarcity, VLA’s performance remains limited. In this work, we introduce Object-Focus Actor (OFA), a novel, data-efficient approach for generalized dexterous manipulation. OFA exploits the consistent end trajectories observed in dexterous manipulation tasks, allowing for efficient policy training. Our method employs a hierarchical pipeline: object perception and pose estimation, pre-manipulation pose arrival and OFA policy execution. This process ensures that the manipulation is focused and efficient, even in varied backgrounds and positional layout. Comprehensive real-world experiments across seven tasks demonstrate that OFA significantly outperforms baseline methods in both positional and background generalization tests. Notably, OFA achieves robust performance with only 10 demonstrations, highlighting its data efficiency. Project website: \href{https://yihanghku.github.io/OFA}{yihanghku.github.io/OFA}.

\end{abstract}

\keywords{Robot manipulation, Imitation learning, Dexterous manipulation} 


\section{Introduction}
Robot manipulation learning from human demonstrations has been considered as a fast way to acquire skills. Generally, dozens to hundreds of trajectories allow to train a policy for a single task \cite{zhao2023learning, fu2024mobile, chi2023diffusion} even for some complex tasks with a considerably high success rate in tests. However, the placement position of objects to be operated are restricted within a limited range and the policies usually take effects in specific scenes. Generalization manipulation, which manipulates objects at an arbitrary position (within the robot's working range) in diverse scenes, is necessary for the robot to be deployed in real-world applications. Inspired by the success of large language model (LLM) \citep{achiam2023gpt, touvron2023llama} and vision language model (VLM) \cite{radford2021learning,li2023blip,chen2024internvl}, where pretraining facilities model to learn general knowledge from large-scale data, many researchers try to find an answer for generalized manipulation using large-scale robot data and a large-scale model \cite{brohan2023rt,kim2024openvla,liu2024rdt, black2024pi}. We call this paradigm as Vision-Language-Action(VLA) in the following text. On the basis of the pretraining model, fine-tuning is conducted for a single specified task. This pipeline requires large-scale robot data, which is expensive at the current stage. Due to the lack of data, the performance of VLA does not show great generalization ability.

To enable practical deployment, robots must generalize their manipulation skills to objects positioned arbitrarily within their workspace and across diverse environments.
In some complex manipulation tasks like using a barcode scanner, dexterous robot hand is a better choice than a two-finger gripper. Generally, manipulation that  requires the use of dexterous robot hand is called dexterous manipulation \cite{okamura2000overview}. Although lots of recent works investigate tele-operation and imitation learning of dexterous manipulation \cite{wang2024dexcap, fu2024humanplus, cheng2024open, ze2024generalizable}, few work discuss the generalization ability of dexterous manipulation. 

\begin{figure*}[t!] 
    \centering
    \includegraphics[width=0.7\textwidth]{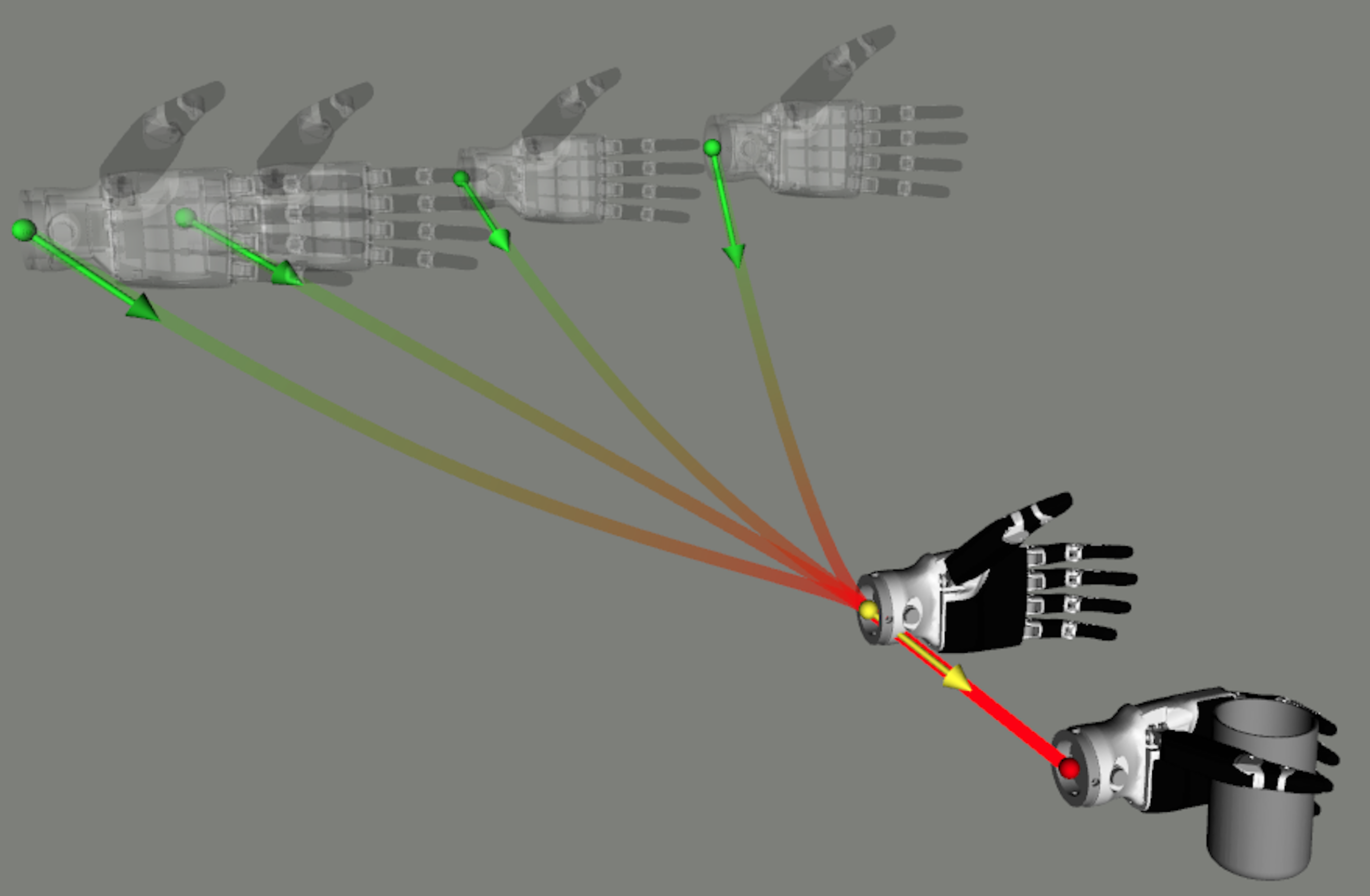}
    \caption{The grasping trajectories of the dexterous robot hand. The four green points denote random initial relative position between the robot hand and the object to be grasped. The colored lines represent the trajectories. The yellow point illustrates the uniform pre-manipulation position and the red point is the final of the trajectories.}
    \label{fig:insight}
    \vspace{-0.2in}
\end{figure*}

The generalization for manipulating objects requires the robot's hand to reach the objects positioned at a random position while remaining within the robot's kinematic and dynamic limits, with an arbitrary hand's initial pose. Once the problem is defined, a straightforward insight is to locate the manipulated object, then generate an available grasp pose and call the trajectory planning algorithm of the robot arm to finish the grasp. This pipeline is sufficiently explored when using a gripper \cite{fang2020graspnet, fang2023anygrasp}, while it is quite challenge to be replicated by a dexterous hand. It is because generating proper grasp poses utilizing multiple fingers is more complicated than using a gripper, most of recent work \cite{li2023gendexgrasp, wang2023dexgraspnet} attempt to solve the problem in simulation while does not transfer to real-world experiments. 

In this work, we propose OFA (Object-Focus Actor) for generalization dexterous manipulation without the requirement of large-scale data. The insight is that when a dexterous manipulation is executed by multiple times, the final part of trajectories are quite similar, regardless of the position of objects to be manipulated. For example, when using a left robot hand to grasp a cup as shown in Fig. \ref{fig:insight}, the hand's starting trajectories vary due to initial relative position's variation (green points in Fig. \ref{fig:insight} denote different initial relative positions). When the robot hand reaches close enough to the object, the trajectories converge to a uniform pose, named pre-manipulation pose and thereafter the final trajectories are same. Not only the wrist motion, but also the trajectories of each hand finger keep consistent in the process. We call this consistent trajectory as object-focus end trajectory. It does not have much variation in each specific task. 

To achieve OFA, we design a hierarchical pipeline. To reach the object focus pre-manipulation pose, the 6-D pose of the target object is calculated using FoundationPose \cite{wen2024foundationpose} firstly. A suitable relative pose which is specified for each object, is then chosen for pre-manipulation pose determination. The pre-manipulation pose arrival from arbitrary initial position is achieved by sample-based trajectory planning method CuRobo \cite{sundaralingam2023curobo}. After reaching the pre-manipulation pose, the robot's hands projecting area in the camera's view are extracted and enlarged to include both the robot hands and the target object. With both the local image and relative pose from the pre-manipulation pose as feedback, the imitation policy begins and finish the manipulation. 

We conduct comprehensive experiments in the real world. Seven tasks vary from target object to operation mode are tested and the results show that our proposed OFA method outperforms baseline method. In positional generalization tests, with object placed beyond data collection area, OFA could keep equivalent performance while the baseline method fails to finish the manipulation. In background generalization tests, OFA method could finish the task even with large background variations. At last, we verify that OFA is a data efficient method that training with 10 demonstrations could achieve a fair enough performance.








\section{Related Work}

\textbf{Imitation learning for dexterous manipulation.} Imitation learning (IL) \citep{fang2019survey} enables robots to learn directly from expert demonstrations. Recently, utilizing IL, many works focus on learning directly from real-world demonstration data obtained by teleportation \citep{wang2024dexcap, fu2024humanplus, cheng2024open, arunachalam2023dexterous, arunachalam2023holo}, for robotic dexterous manipulation. Specifically, DexCap \citep{wang2024dexcap} employs a point cloud-based diffusion policy to seamlessly replicate human actions with robot hands. 
HumanPlus \citep{fu2024humanplus} uses a decoder-only transformer network to perform supervised behavior cloning to train a skill policy and predict future image features to prevent the policy from overfitting to proprioception. However, in these works, the positions of the manipulated objects are relatively fixed, and the issue of generalizing object positions, especially with limited data, has not been tackled. Our work aims to tackle the challenge of object position generalization in robotic dexterous manipulation by learning a unified core manipulation trajectory.

\textbf{Object-centric method for robot manipulation.} Object visual representations are widely used for robot manipulation to improve success rate and generalization through pre-trained vision models \citep{zhu2023learning,zhu2024vision,valassakis2022demonstrate, huang2024rekep, kerr2024robot}. These works usually extract object mask \citep{zhu2023learning, valassakis2022demonstrate}, 6D pose \cite{pan2025omnimanip}, keypoints \cite{zhu2024vision, huang2024rekep, zhou2024code} to provide object prior for policy learning, which could prevent the policy itself from handling complex image observation especially in real-world scenarios. GROOT \citep{zhu2023learning} utilizes object segmentation to extract masks mainly for improving the manipulation success rate against changed background. ORIGIN \cite{zhu2024vision} designs a graph-based object-centric representation method that further generalizes to various object positional layout. However, these works only do image processing to focus on the object to be operated, while does not change the control policy to an object-centric paradigm. Our work aims to achieve a full object-centric manipulation method that consider both object-centric image processing and object-centric control policy.

\textbf{Quantity of demonstration for manipulation policy learning.} Expert demonstrations are crucial for manipulation policy learning. VINN \cite{pari2021surprising} learns a policy for Door Openning, Stacking and Pushing using 71 demonstrations. ALOHA \cite{zhao2023learning} and Mobile-ALOHA \cite{fu2024mobile} propose the method ACT that learns a policy for desk manipulation tasks from 50 demonstrations. Diffusion Policy \cite{chi2023diffusion} trains for four realworld tasks that require 90 demonstrations at least for each task. Although these works have achieved good performance in the specific tasks and layout, the rollout experiments barely consider the generalization ability of the learned policy. In \cite{lin2024data}, the authors discuss about the data scaling law for imitation learning, which conduct experiments on various enviorments to test the manipulation performance. They conclude that each task is recommandded to collect 50 demonstrations in each environment and more than 1000 demonstrations in various environments. In recent VLA methods \citep{kim2024openvla, black2024pi, liu2024rdt}, the authors use large-scale robot data to pretrain the model and then fine-tune the model on the specific task. 
Our work aims to find the core learnable trajectory in a manipulation and tries to exploit the demonstrations more efficiently. Therefore, the policy can be trained with less demonstrations and still achieve good generalization ability.

\vspace{-2.0mm}
\section{Method}
In this section, the Object-Focus Actor (OFA) method is formally proposed. First, the overall design of OFA is outlined. Then, detailed explanations of manipulating-object perception and pose estimation are given. Next, the pre-manipulation pose arrival module is introduced. Finally, an in-depth description of object-focus policy learning is presented.
\subsection{Overall Design of OFA}

\begin{figure*}[htp!] 
    \centering
    \includegraphics[width=1.0\textwidth]{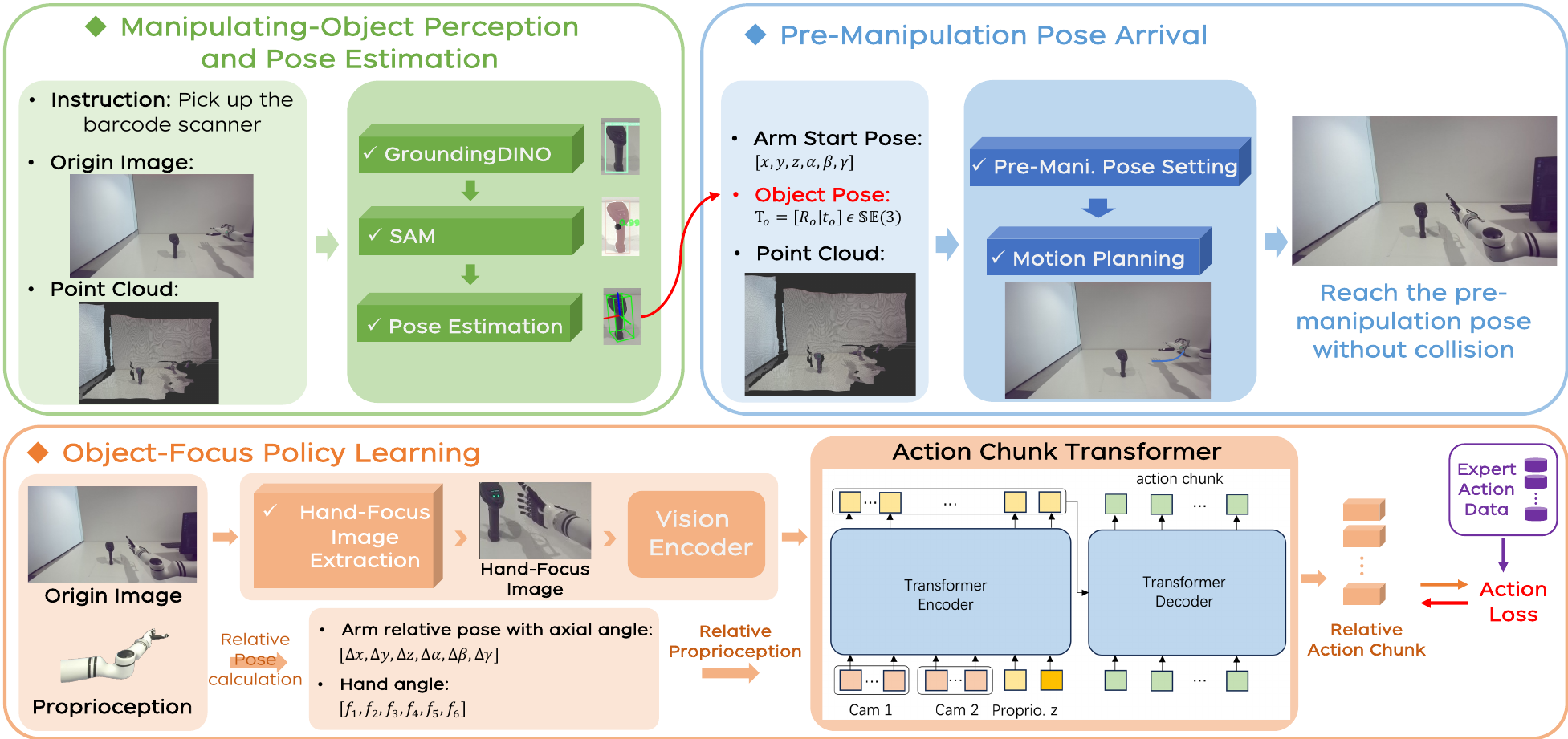}
    \caption{Overall design of OFA}
    \label{fig:overall_method}
\end{figure*}

As shown in Figure \ref{fig:overall_method}, the overall structure of the proposed OFA mainly consists of the following three modules: 1) manipulating-object perception and pose estimation, 2) pre-manipulation pose arrival, 3) object-focus policy learning. Specifically, OFA begins by identifying the name of the object to be manipulated based on user instructions, such as a cup. Subsequently, using original image and point cloud data, the pose of the manipulating object is extracted through perception and pose estimation algorithm modules, e.g., GroundingDINO \citep{liu2023grounding}, SAM \cite{kirillov2023segment}, FoundationPose \cite{wen2024foundationpose}, etc. Based on the object’s pose, the pre-manipulation pose of the robotic hand is set, and a motion planning algorithm (i.e., cuRobo \citep{sundaralingam2023curobo}) is employed to guide the robotic hand to the pre-manipulation pose. Finally, object-focus policy learning is designed to enable the robotic hand to learn and execute the final dexterous manipulation, such as picking up the cup. Specifically, OFA is designed to predict core manipulation trajectories utilizing hand-focus images. Furthermore, in order to enhance the model's generalization ability for the position of objects, the models use relative proprioception information and relative action chunks.

\vspace{-0.3cm}

\subsection{Manipulating-Object Perception and Pose Estimation}

\textbf{Object locating:} To conveniently extract the interested object from origin image, the open-set grounding object detector, GroundingDINO \cite{liu2023grounding} is exploited. It passes the object prior to subsequent segmentation model, SAM \cite{kirillov2023segment} for pixel-level location of the object. 

\textbf{Pose estimation:} The pre-manipulation pose of robot hand depends on a relative displacement and rotation from the object. Therefore, it is crucial to estimate the 6D pose of the object, precisely and consistently. FoundationPose \cite{wen2024foundationpose} is a unified pose estimation method. It adopts either multi-views images of the object for model-free pose estimation or single-view image together with the CAD model for model-based pose estimation. In the circumstance of robot manipulation, only a single-view image is provided, thus the model-based way is utilized. 

For a manipulation task, the manipulating object's 3D models are firstly constructed by 3D scanner. In inference stage, a RGB image, an aligned depth image is collected firstly. Then a mask image which segments the target object is generated from the last step. Together with the CAD model, all the images are sent to the pose estimation model to estimate the object's 6D pose. 

\subsection{Pre-Manipulation Pose Arrival}


\textbf{Pre-manipulation pose setting for robotic hand:} For a robotic hand, an object may have multiple suitable pre-manipulation poses. For instance, a cylindrical cup can be grasped from the left, right, front, or back. Assigning a unique pre-manipulation pose to each individual object can be inefficient and labor-intensive. Through observation, we found that it is possible to categorize objects and extract a unified pre-manipulation pose for each category, thereby eliminating the need to configure a pose for each specific object. Intuitively, the way a robotic hand manipulates an object is closely linked to the object’s shape and size. For example, a cylindrical cup can be grasped from the right side with the right hand, while a cup with a handle requires grasping the handle, and a pen should be grasped from above. Therefore, we can categorize objects based on their shape and size and set a unified pre-manipulation pose for each category. 



We represent the 6D pose of a manipulated object as $\mathbf{T}_{o} = [\mathbf{R}_o | \mathbf{t}_o] \in \mathbb{SE}(3)$, where $\mathbf{R}_o \in \mathbb{SO}(3)$ is rotation matrix representing the orientation and $\mathbf{t}_o \in  \mathbb{R}^3$ is the translation.

The pre-manipulation pose has a static orientation offset $\Delta \mathbf{R}_0 \in \mathbb{SO}(3)$ and translation offset $\Delta \mathbf{t}_o \in \mathbb{R}^3$ from $\mathbf{T}_{o}$. Therefore, the pre-manipulation pose is calculated:
\small
\begin{equation}
    \label{eq_pre_manipulation}
    \mathbf{T}_{m} = [\mathbf{R}_m | \mathbf{t}_m] = [\mathbf{R}_o \oplus \Delta \mathbf{R}| \mathbf{t}_o + \Delta \mathbf{t}]
\end{equation}
where $\oplus$ denotes the update on $\mathbb{SO}(3)$.

\textbf{Motion planning:} After obtaining the pre-manipulation pose of the robotic hand, it is necessary for the robotic hand to reach this pose without collisions. In our paper, the CuRobo \citep{sundaralingam2023curobo} planner from NVIDIA is utilized to generate collision-free motion trajectories for the robotic hand. This planner employs a multi-GPU parallel optimization approach to accelerate the generation of motion trajectories. In particular, a large number of spheres are used as the collision representation for the robot arm and hand. Subsequently, given the target pose for pre-manipulation and the real-time point cloud information of the environment, CuRobo can generate a collision-free path for the robotic hand.


\subsection{Object-Focus Policy Learning}

\begin{figure}[htp]
    \centering
    \includegraphics[width=1.0\textwidth]{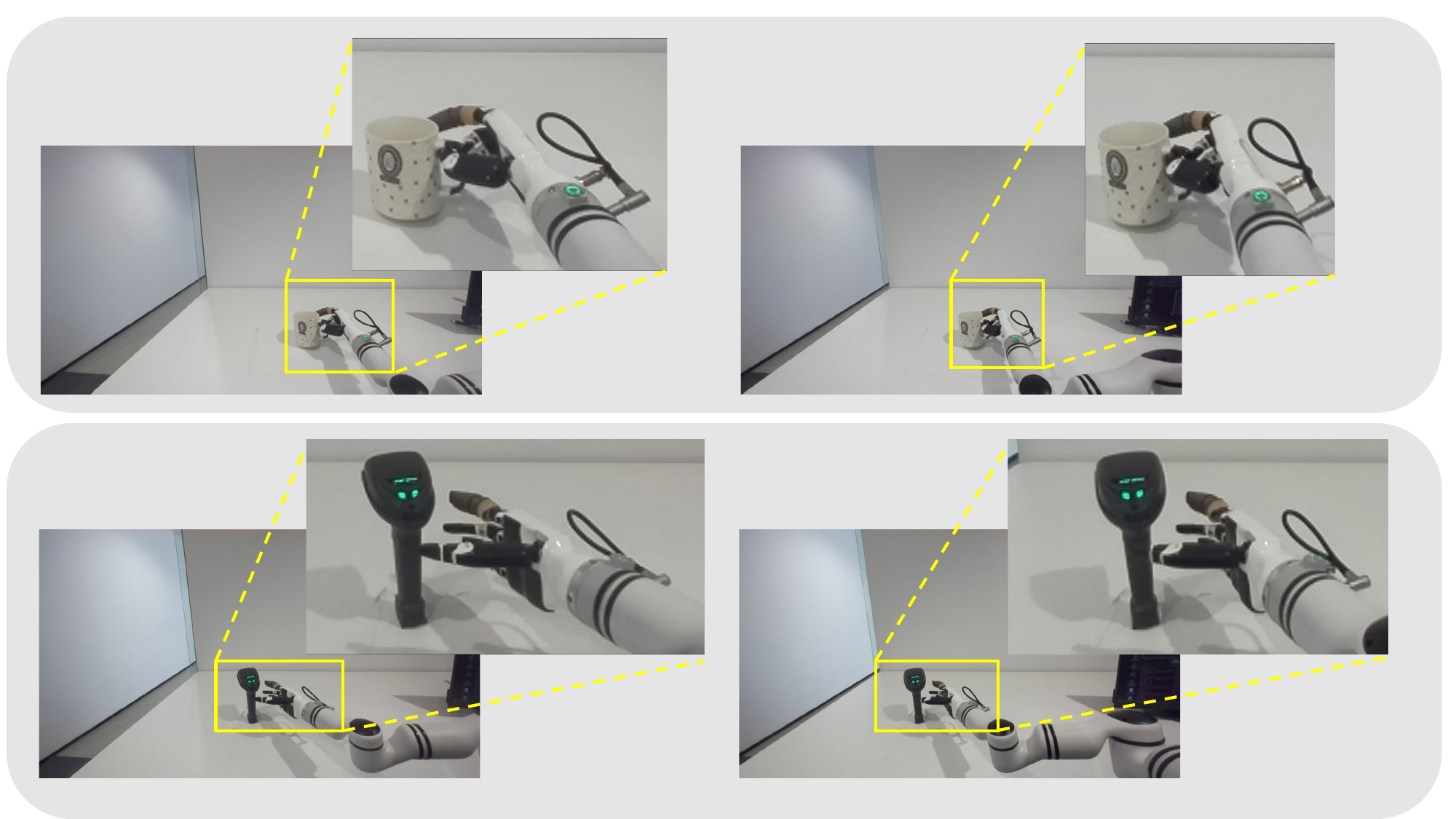}
    \caption{The extraction of robot hand focus area. The left image and the right image are captured by the pair RGB stereo camera respectively. In each task, the hand focus area almost only includes the dexterous hand and the target object.}
    \label{fig:object-focus}
\end{figure}

\textbf{Hand-Focus images}
The Hand-Focus images are constructed using the forward dynamics of robot arm and hand. The pose of wrist in arm's base coordinates ${}^{a}\mathbf{T}_{w} = [{}^{a}\mathbf{R}_{w}| {}^{a}\mathbf{t}_{w}]$ are obtained with the proprioception of robot arm. The finger joints $\mathbf{J}_h = [j_1, j_2,\cdots,j_6]$ are similarly obtained from the robot hand's feedback. A full digital robot hand is built according to the robot hand's URDF file and proprioceptions ${}^{a}\mathbf{T}_{w}$, $\mathbf{J}_h$. 
Then it is transformed to camera coordinate with the extrinsic ${}^{e}\mathbf{T}_{a}$ :  ${}^{e}\mathbf{T}_{w} = {}^{e}\mathbf{T}_{a} {}^{a}\mathbf{T}_{w}$. Then the 3D points of the robot hand are projected to camera image and the hand's area is represented by the enclosing rectangle of all the projected points. As discussed before, the pre-manipulation point is close to the object to be manipulated, thus we enlarge the hand's area to its twice as shown in fig. \ref{fig:object-focus} and allow the object include in the local area. Finally, this area is resized to a unified size as the resulted Hand-Focus image.

\textbf{Relative proprioception:} The relative proprioception of a robot hand consists of relative pose $\mathbf{T}_{rel} = [\Delta \mathbf{p}, \Delta \bm{\omega}]$ and finger angle $\mathbf{J}_h = [j_1, j_2,\cdots,j_6]$. Specifically, $\mathbf{T}_{rel}$ is the pose relative to the pre-manipulation pose, where $\Delta \mathbf{p}$ denotes the relative value of the position, $\Delta \bm{\omega}$ represents the relative orientation in the format of axis angle.

\textbf{Relative action chunk:} In this paper, to mitigate compounding errors and the causal confusion issue in imitation learning \citep{de2019causal}, the policy is trained to predict an action chunk instead of a single action. Specifically, to allow the model to efficiently learn positional generalization, the training action data employ the relative action chunk $\hat{a}_{t:t+k} = [\hat{a}_{t},...,\hat{a}_{t+k}]$, where $\hat{a}_{t} = [p_{rel}^{a}, f^{a}]$ denotes the relative action of a robot hand at time $t$, $p_{rel}^{a}$ indicates the relative pose of the robot action at time $t$, expressed in the axis angle format, relative to the pre-manipulation pose, $f^{a}$ represents the finger angles of the robot hand.

\textbf{Model architecture:} As shown in Figure \ref{fig:overall_method}, similar to Action Chunking with Transformers (ACT) \citep{zhao2023learning}, a hand-focus model is built as a conditional variational autoencoder (CVAE) \citep{sohn2015learning}, to generate relative action chunks based on current hand-focus observation $o_{t} = [Images, \mathbf{T}_{rel}, \mathbf{J}_h]$. 

\textbf{Training loss:} Referring to CVAE, the training loss of this model is defined as follows,
\begin{equation}
    \label{hand_focus_model_training_loss}
    \mathcal{L}_{hand\_focus} = MSE(a_{t:t+k},\hat{a}_{t:t+k}) + \eta D_{KL}(q(z|\hat{a}_{t:t+k},\bar{o}_{t})\,\|\,\mathcal{N}(0,I)),
\end{equation}
where the first term on the left side of the equation is the reconstruction loss, the second term regularizes the encoder to a Gaussian prior, $\eta$ is a hyperparameter, and $\bar{o}_{t}$ represents $o_{t}$ without image observations.

In addition, the training and inference pseudocode are given in Appendix \ref{sec:appendixB}.



\vspace{-2.0mm}
\section{Experiments}
We conduct quantitative evaluations in the real world to validate our OFA method. Firstly, Section \ref{experimental_setting} outlines the experimental setup, detailing the hardware used for testing, the seven tasks chosen to encompass a wide range of daily manipulation behaviors for assessing the effectiveness of our proposed method against the baseline, and the data collection process for imitation learning. Then, the quantitative results are shown in Sec. \ref{sec_experiment:effectiveness}. The generalization abilities towards different object position and various object background are further evaluated in Sec. \ref{sec_experiment:generalization}. At Sec. \ref{sec_experiment:quantity}, the requirement of demonstration amount is investigated.

\subsection{Experimental Setting}
\label{experimental_setting}

\subsubsection{Hardware}
\label{sec_experiment: hardware}
\begin{figure}
    \centering
    \includegraphics[width=1.0\linewidth]{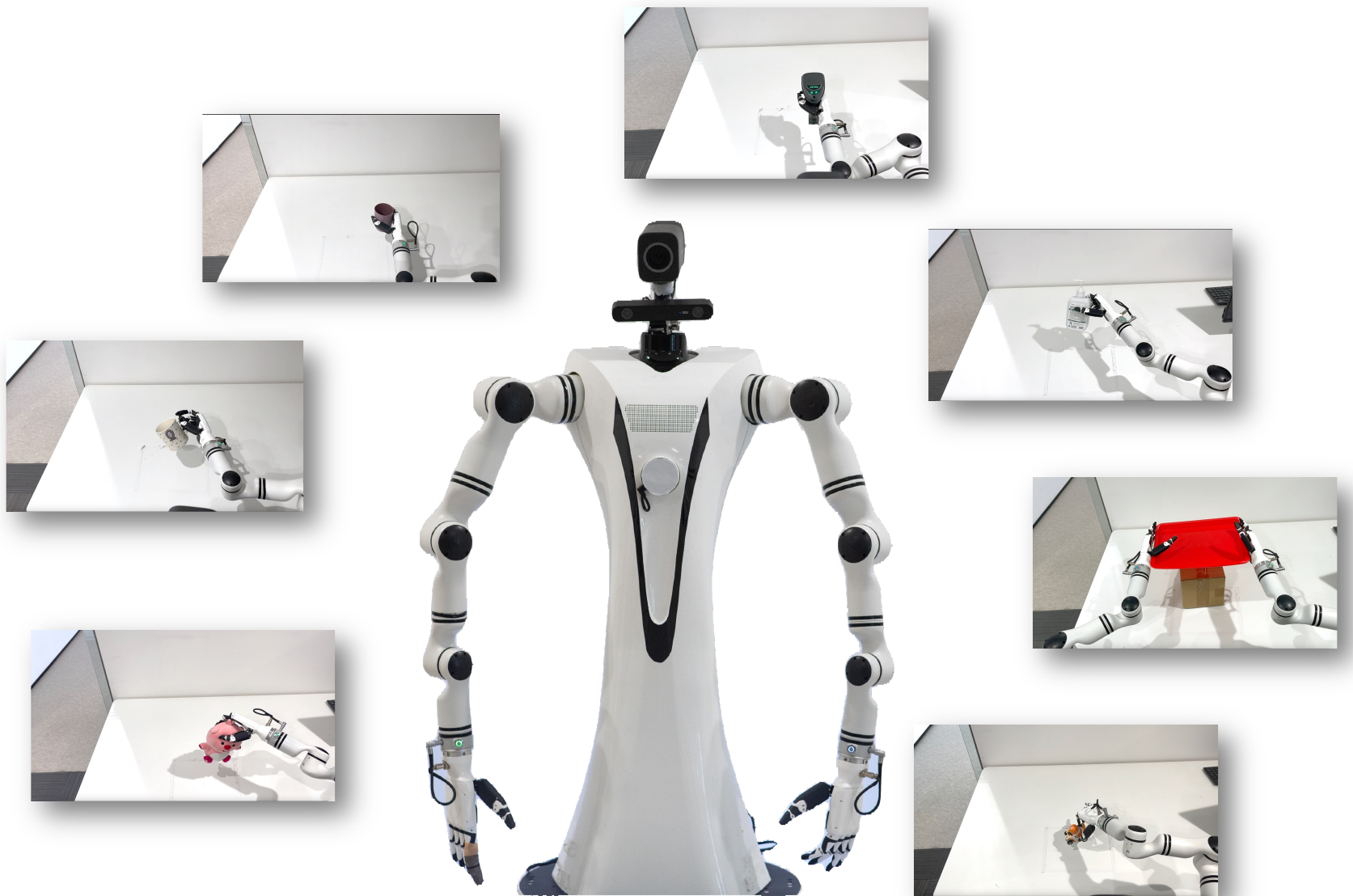}
    \caption{The bimanual robot used in our experiment. The mainly used modules for this paper are Realman RM65-B 6-DoF arms, Inspire RH56BFX dexterous hands and a ZED2 camera installed on the head.}
    \label{fig:hardware}
\end{figure}

Our experiments are conducted on the Realman Embodied Dual-arm platform as shown in fig. \ref{fig:hardware}, equipped with two 6 DoF robotic arms and a mobile vehicle. This work does not involve mobile manipulation, thus the mobile ability of this platform is not shown. Dexterous robot hands form Inspire Company which has 6 active joint are installed as the end effectors. The stereo camera ZED2 which has $110^{\degree} (H) \times 70^{\degree} (V)$ FoV is installed on the head to capture RGB images and process for depth information. 

\subsubsection{Task Setting}
\label{sec_experiment: task setting}

In our experiments, to comprehensively evaluate the effectiveness of the proposed method, $7$ challenging dexterous manipulation tasks are designed, including $6$ single-hand tasks and $1$ dual-hand tasks. The single-hand tasks include: grasping a cup, taking a mug, holding a barcode scanner, catching a loopy, pinching a toy, and grasping a sanitizer. The dual-hand tasks include lifting a tray with both hands. Detailed descriptions of these tasks are shown in Appedndix \ref{sec:appendixA}.

\subsubsection{Data Collection}

For the seven tasks described above, we need to collect human demonstration data for imitation learning of the policy. Specifically, we used strain gauge gloves and a positioning camera as teleoperation devices to control a dexterous hand for data collection. During the collection of demonstration data for each task, the position of the manipulated objects is not fixed but is randomly placed within a rectangular area. Additionally, the initial position of the dexterous hand is fixed.

\subsection{Quantitative Experiment Results}
\label{sec_experiment:effectiveness}
To verify the effectiveness of the proposed method, different methods and their variants are evaluated on these challenging tasks, as follows.

\begin{itemize}
    \item ACT \citep{zhao2023learning}: Action Chunking with Transformers (ACT) model.
    \item OFA w/o rel: A variant of the proposed OFA using absolute pose as proprioception observation.
    \item OFA w/o of: A variant of the proposed OFA using original images.
    \item OFA w/o rel-of: A variant of the proposed OFA that does not use relative pose or Hand-Focus image.
    \item OFA: Our proposed Object-Focus Actor that use both relative pose and Hand-Focus image.

\end{itemize}

The different methods are evaluated on the seven tasks mentioned above. The results are shown in Table \ref{performance_of_methods}. As we expected, the \textit{OFA} method demonstrates superior performance across all tasks compared to other methods. In contrast, the \textit{ACT} method fails due to the random placement of objects within a certain area. With limited data, the \textit{ACT} method struggles to handle this randomness, as it is particularly sensitive to object positions. 

Additionally, by comparing the \textit{OFA w/o rel} method with the \textit{OFA} method, we find that using relative poses as proprioception and action chunks significantly benefits the method. This is because learning relative poses reduces the action space that needs to be learned, concentrating only on a unified operational trajectory. Conversely, by comparing the \textit{OFA w/o of} method with the \textit{OFA} method, we discover that extracting hand-focus images is not always beneficial, such as in tasks like grasping a cup or holding a scanner. This is because our experimental environment is relatively clean or ideal, reducing the impact of hand-focus images. Further background generalization tests will analyze the role of hand-focus images. In summary, these experimental results thoroughly validate the effectiveness of the proposed method, regardless of whether the tasks involve single-handed or bimanual dexterous manipulation.


\begin{table}[t]
\small
\centering
\caption{Success rate ($\%$) of the comparison methods using $30$ human demonstrations. The results are obtained with $10$ evaluations.}
\setlength\tabcolsep{1.5pt} 
\begin{tabular}{@{}c|c|c|c|c|c@{}}
\toprule
Task & ACT & OFA w/o rel-of & OFA w/o rel & OFA w/o of  & \textbf{OFA}  \\\midrule
Grasp Cup & 20 & 40 & 30 & \textbf{90} & \textbf{90} \\
Take Mug & 10 & 30 & 20 & 40 & \textbf{60} \\
Hold Scanner & 30 & 50 & 30 & \textbf{90} & \textbf{90} \\
Catch Loopy & 40 & 40 & 70 & \textbf{90}  & 80 \\
Pinch Toy & 20 & \textbf{40} & 10 & 30  & \textbf{40} \\
Grasp Sanitizer & 30 & 70 & 50 & 80 & \textbf{100} \\
Lift Tray & 10 & \textbf{90} & 60 & 60 & \textbf{100} \\\bottomrule
\end{tabular}
\label{performance_of_methods}
\vspace{-0.2cm}
\end{table}

\subsection{Generalization Analysis}
\label{sec_experiment:generalization}
Beyond evaluating the effectiveness of the methods, we also analyze their generalization to different positions of objects and background variations.

\begin{figure*}[htbp]
    \centering
    \subfigure[Success rate ($\%$) for different manipulating-object position.]{
        \includegraphics[width=0.48\textwidth]{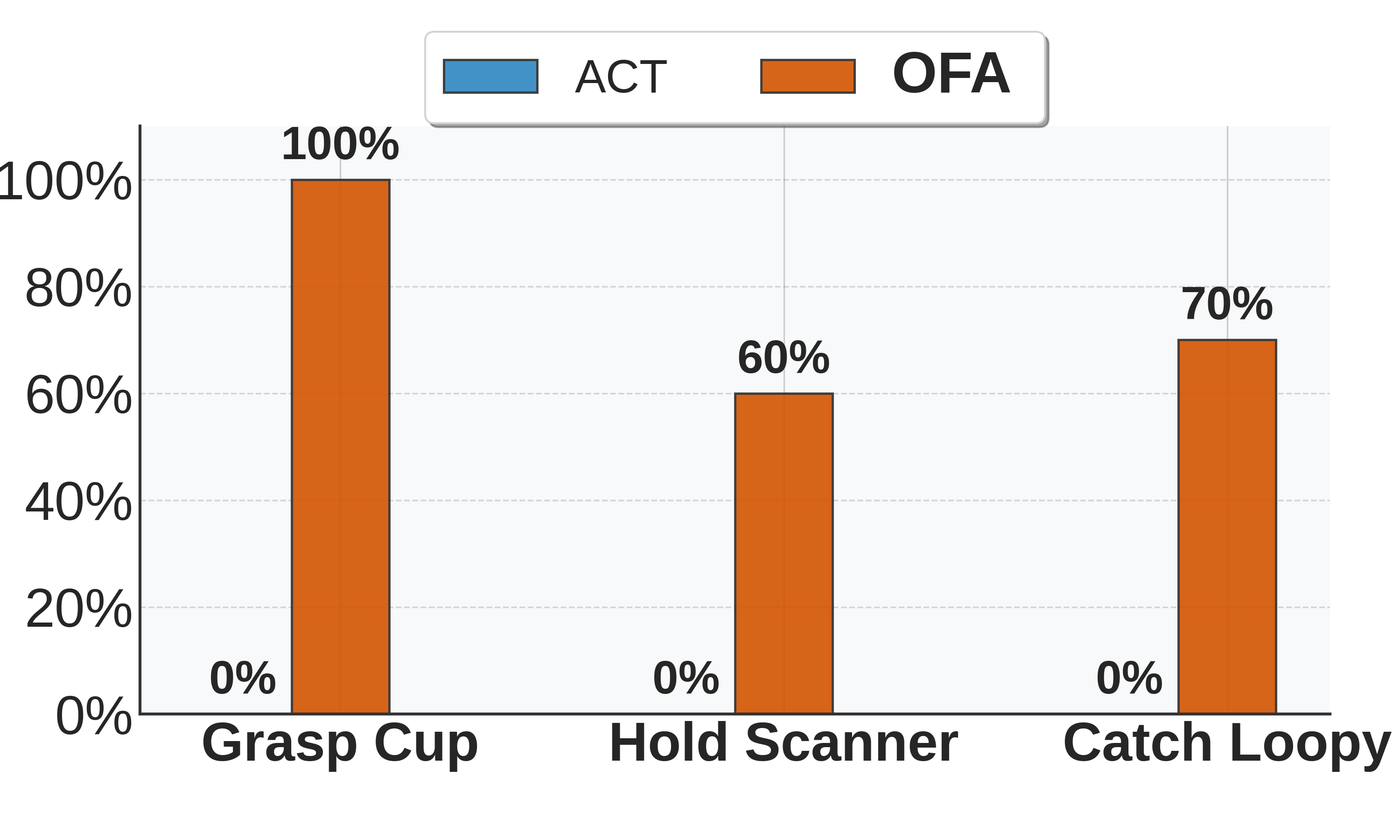}
    }
    \subfigure[Success rate ($\%$) for background variations.]{
        \includegraphics[width=0.48\textwidth]{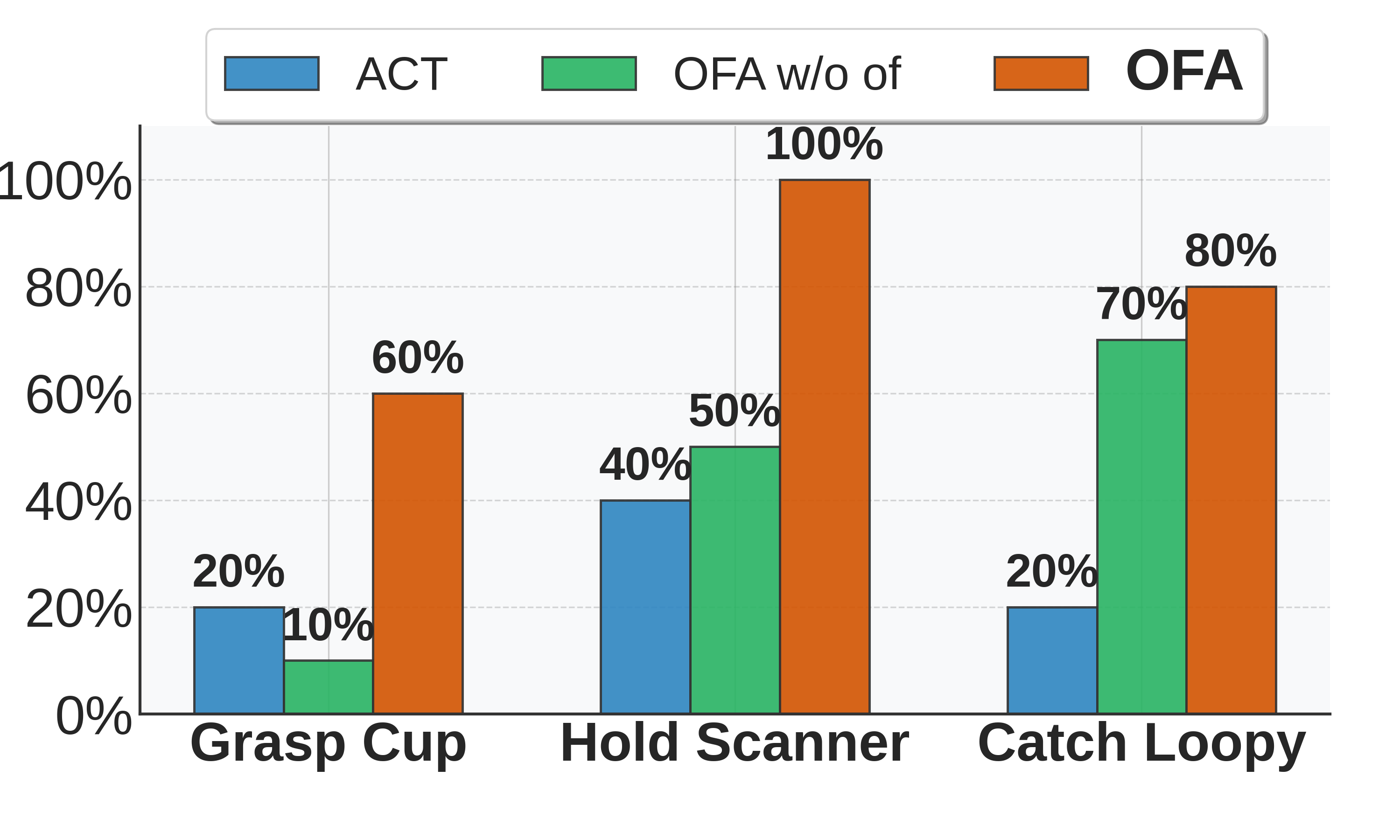}
    }
    \caption{Generalization testing of different methods.}
    \label{fig:generalization testing}
    \vspace{-0.2in}
\end{figure*}

\textbf{Different manipulating-object position:} We place the manipulated objects in positions different from those in the training dataset and conduct 10 tests for \textit{ACT} and \textit{OFA} in three tasks: gripping the cup, holding the scanner, and catching loopy. The test results are presented in Figure \ref{fig:generalization testing}-(a). The results indicate that the proposed \textit{OFA} method exhibits strong positional generalization capabilities, maintaining a high success rate even for object positions that have never been encountered before. In contrast, the \textit{ACT} method completely fails, with a success rate of zero. This outcome is expected, as the \textit{ACT} method performs poorly even with previously encountered object positions, let alone new ones.



\textbf{Background variations:} We alter the background environment to evaluate the robustness of the methods against background interference, conducting 10 tests for \textit{ACT}, \textit{OFA w/o of}, and \textit{OFA} in the three tasks: gripping the cup, holding scanner, and catching loopy. The results are shown in Figure  \ref{fig:generalization testing}-(b). The experimental results demonstrate that the proposed \textit{OFA} method possesses the strongest background generalization capability compared to the \textit{ACT} and \textit{OFA w/o of} methods. In contrast, the \textit{ACT} method performs poorly due to its lack of focus on the core operational area, making it susceptible to environmental background influences. Additionally, by comparing the experimental results of \textit{OFA w/o of} and \textit{OFA}, we further investigate the effectiveness of hand-focus images. Particularly in tasks involving smaller objects, such as the Grasp Cup task, \textit{OFA w/o of} achieves only a $10\%$ success rate due to the absence of images focused on the manipulation area, whereas \textit{OFA} achieves a $60\%$ success rate.



\subsection{Sensitivity analysis of the quantity of demonstration data}
\label{sec_experiment:quantity}
Besides, we further analyze the sensitivity to the quantity of demonstration data. We select three representative tasks: Grasp Cup, Hold Scanner, and Grasp Sanitizer, for this sensitivity analysis. Both the \textit{ACT} and \textit{OFA} methods are trained using $10$, $20$, and $30$ demonstrations for each task and then evaluate $10$ times. The results are shown in Table \ref{performance_of_methods_data_efficient}. The findings indicate that the \textit{OFA} method exhibits strong data efficiency compared to the \textit{ACT} method. Specifically, the performance of \textit{OFA} with just $10$ demonstrations surpasses that of the \textit{ACT} method with $30$ demonstrations. Moreover, for the proposed \textit{OFA} method, $30$ demonstrations are sufficient to achieve high performance, and for some tasks, such as holding a scanner, even $20$ demonstrations are adequate. Compared to \textit{ACT}, the proposed \textit{OFA} method shows less sensitivity to the quantity of data required, further underscoring its data-efficient capabilities.


\begin{table}[t]
\small
\centering
\caption{Success rate ($\%$) of the methods with different amounts of demonstration data.}
\begin{tabular}{@{}c|cc|cc|cc@{}}
\toprule
\multicolumn{1}{c}{\multirow{2}{*}{Task}} & \multicolumn{2}{c}{10 Demo.} & \multicolumn{2}{c}{20 Demo.} & \multicolumn{2}{c}{30 Demo.} \\\cmidrule(lr){2-7}
\multicolumn{1}{c}{} & \multicolumn{1}{c}{ACT} & \multicolumn{1}{c}{\textbf{OFA}} & \multicolumn{1}{c}{ACT} & \textbf{OFA} & \multicolumn{1}{c}{ACT} & \textbf{OFA}  \\\cmidrule(lr){1-7}
Grasp Cup & 30 & 40 & 30 & 80 & 20 & \textbf{90}  \\
Hold Scanner & 10 & 70 & 60 & \textbf{90} & 60 & \textbf{90}  \\
Grasp Sanitizer & 10 & 60 & 10 & 70 & 30 & \textbf{100}  \\
\bottomrule
\end{tabular}
\label{performance_of_methods_data_efficient}
\vspace{-0.2cm}
\end{table}


\vspace{-2.0mm}
\section{Conclusion}
\vspace{-1.0mm}
In this paper, we present the Object-Focus Actor (OFA), a novel and data-efficient approach to generalized dexterous manipulation. Our method demonstrates a substantial improvement over the baseline, exceeding performance by more than 50\% in in-distribution real-world experiments. Furthermore, OFA showcases robust generalization capabilities in out-of-distribution scenarios, effectively handling challenges such as positional offsets and background variations. Notably, OFA maintains commendable performance with as few as 10 demonstrations, underscoring its exceptional data efficiency. These results highlight OFA’s potential as a powerful tool for advancing dexterous manipulation in diverse and dynamic environments.


\bibliography{references}  

\begin{thebibliography}{39}
\providecommand{\natexlab}[1]{#1}
\providecommand{\url}[1]{\texttt{#1}}
\expandafter\ifx\csname urlstyle\endcsname\relax
  \providecommand{\doi}[1]{doi: #1}\else
  \providecommand{\doi}{doi: \begingroup \urlstyle{rm}\Url}\fi

\bibitem[Zhao et~al.(2023)Zhao, Kumar, Levine, and Finn]{zhao2023learning}
T.~Z. Zhao, V.~Kumar, S.~Levine, and C.~Finn.
\newblock Learning fine-grained bimanual manipulation with low-cost hardware.
\newblock \emph{arXiv preprint arXiv:2304.13705}, 2023.

\bibitem[Fu et~al.(2024)Fu, Zhao, and Finn]{fu2024mobile}
Z.~Fu, T.~Z. Zhao, and C.~Finn.
\newblock Mobile aloha: Learning bimanual mobile manipulation with low-cost
  whole-body teleoperation.
\newblock \emph{arXiv preprint arXiv:2401.02117}, 2024.

\bibitem[Chi et~al.(2023)Chi, Xu, Feng, Cousineau, Du, Burchfiel, Tedrake, and
  Song]{chi2023diffusion}
C.~Chi, Z.~Xu, S.~Feng, E.~Cousineau, Y.~Du, B.~Burchfiel, R.~Tedrake, and
  S.~Song.
\newblock Diffusion policy: Visuomotor policy learning via action diffusion.
\newblock \emph{The International Journal of Robotics Research}, page
  02783649241273668, 2023.

\bibitem[Achiam et~al.(2023)Achiam, Adler, Agarwal, Ahmad, Akkaya, Aleman,
  Almeida, Altenschmidt, Altman, Anadkat, et~al.]{achiam2023gpt}
J.~Achiam, S.~Adler, S.~Agarwal, L.~Ahmad, I.~Akkaya, F.~L. Aleman, D.~Almeida,
  J.~Altenschmidt, S.~Altman, S.~Anadkat, et~al.
\newblock Gpt-4 technical report.
\newblock \emph{arXiv preprint arXiv:2303.08774}, 2023.

\bibitem[Touvron et~al.(2023)Touvron, Lavril, Izacard, Martinet, Lachaux,
  Lacroix, Rozi{\`e}re, Goyal, Hambro, Azhar, et~al.]{touvron2023llama}
H.~Touvron, T.~Lavril, G.~Izacard, X.~Martinet, M.-A. Lachaux, T.~Lacroix,
  B.~Rozi{\`e}re, N.~Goyal, E.~Hambro, F.~Azhar, et~al.
\newblock Llama: Open and efficient foundation language models.
\newblock \emph{arXiv preprint arXiv:2302.13971}, 2023.

\bibitem[Radford et~al.(2021)Radford, Kim, Hallacy, Ramesh, Goh, Agarwal,
  Sastry, Askell, Mishkin, Clark, et~al.]{radford2021learning}
A.~Radford, J.~W. Kim, C.~Hallacy, A.~Ramesh, G.~Goh, S.~Agarwal, G.~Sastry,
  A.~Askell, P.~Mishkin, J.~Clark, et~al.
\newblock Learning transferable visual models from natural language
  supervision.
\newblock In \emph{International conference on machine learning}, pages
  8748--8763. PMLR, 2021.

\bibitem[Li et~al.(2023)Li, Li, Savarese, and Hoi]{li2023blip}
J.~Li, D.~Li, S.~Savarese, and S.~Hoi.
\newblock Blip-2: Bootstrapping language-image pre-training with frozen image
  encoders and large language models.
\newblock In \emph{International conference on machine learning}, pages
  19730--19742. PMLR, 2023.

\bibitem[Chen et~al.(2024)Chen, Wu, Wang, Su, Chen, Xing, Zhong, Zhang, Zhu,
  Lu, et~al.]{chen2024internvl}
Z.~Chen, J.~Wu, W.~Wang, W.~Su, G.~Chen, S.~Xing, M.~Zhong, Q.~Zhang, X.~Zhu,
  L.~Lu, et~al.
\newblock Internvl: Scaling up vision foundation models and aligning for
  generic visual-linguistic tasks.
\newblock In \emph{Proceedings of the IEEE/CVF Conference on Computer Vision
  and Pattern Recognition}, pages 24185--24198, 2024.

\bibitem[Brohan et~al.(2023)Brohan, Brown, Carbajal, Chebotar, Chen,
  Choromanski, Ding, Driess, Dubey, Finn, et~al.]{brohan2023rt}
A.~Brohan, N.~Brown, J.~Carbajal, Y.~Chebotar, X.~Chen, K.~Choromanski,
  T.~Ding, D.~Driess, A.~Dubey, C.~Finn, et~al.
\newblock Rt-2: Vision-language-action models transfer web knowledge to robotic
  control.
\newblock \emph{arXiv preprint arXiv:2307.15818}, 2023.

\bibitem[Kim et~al.(2024)Kim, Pertsch, Karamcheti, Xiao, Balakrishna, Nair,
  Rafailov, Foster, Lam, Sanketi, et~al.]{kim2024openvla}
M.~J. Kim, K.~Pertsch, S.~Karamcheti, T.~Xiao, A.~Balakrishna, S.~Nair,
  R.~Rafailov, E.~Foster, G.~Lam, P.~Sanketi, et~al.
\newblock Openvla: An open-source vision-language-action model.
\newblock \emph{arXiv preprint arXiv:2406.09246}, 2024.

\bibitem[Liu et~al.(2024)Liu, Wu, Li, Tan, Chen, Wang, Xu, Su, and
  Zhu]{liu2024rdt}
S.~Liu, L.~Wu, B.~Li, H.~Tan, H.~Chen, Z.~Wang, K.~Xu, H.~Su, and J.~Zhu.
\newblock Rdt-1b: a diffusion foundation model for bimanual manipulation.
\newblock \emph{arXiv preprint arXiv:2410.07864}, 2024.

\bibitem[Black et~al.(2024)Black, Brown, Driess, Esmail, Equi, Finn, Fusai,
  Groom, Hausman, Ichter, et~al.]{black2024pi}
K.~Black, N.~Brown, D.~Driess, A.~Esmail, M.~Equi, C.~Finn, N.~Fusai, L.~Groom,
  K.~Hausman, B.~Ichter, et~al.
\newblock $\pi_0$:a vision-language-action flow model for general robot
  control.
\newblock \emph{arXiv preprint arXiv:2410.24164}, 2024.

\bibitem[Okamura et~al.(2000)Okamura, Smaby, and Cutkosky]{okamura2000overview}
A.~M. Okamura, N.~Smaby, and M.~R. Cutkosky.
\newblock An overview of dexterous manipulation.
\newblock In \emph{Proceedings 2000 ICRA. Millennium Conference. IEEE
  International Conference on Robotics and Automation. Symposia Proceedings
  (Cat. No. 00CH37065)}, volume~1, pages 255--262. IEEE, 2000.

\bibitem[Wang et~al.(2024)Wang, Shi, Wang, Zhang, Fei-Fei, and
  Liu]{wang2024dexcap}
C.~Wang, H.~Shi, W.~Wang, R.~Zhang, L.~Fei-Fei, and C.~K. Liu.
\newblock Dexcap: Scalable and portable mocap data collection system for
  dexterous manipulation.
\newblock \emph{arXiv preprint arXiv:2403.07788}, 2024.

\bibitem[Fu et~al.(2024)Fu, Zhao, Wu, Wetzstein, and Finn]{fu2024humanplus}
Z.~Fu, Q.~Zhao, Q.~Wu, G.~Wetzstein, and C.~Finn.
\newblock Humanplus: Humanoid shadowing and imitation from humans.
\newblock \emph{arXiv preprint arXiv:2406.10454}, 2024.

\bibitem[Cheng et~al.(2024)Cheng, Li, Yang, Yang, and Wang]{cheng2024open}
X.~Cheng, J.~Li, S.~Yang, G.~Yang, and X.~Wang.
\newblock Open-television: Teleoperation with immersive active visual feedback.
\newblock \emph{arXiv preprint arXiv:2407.01512}, 2024.

\bibitem[Ze et~al.(2024)Ze, Chen, Wang, Chen, He, Yuan, Peng, and
  Wu]{ze2024generalizable}
Y.~Ze, Z.~Chen, W.~Wang, T.~Chen, X.~He, Y.~Yuan, X.~B. Peng, and J.~Wu.
\newblock Generalizable humanoid manipulation with improved 3d diffusion
  policies.
\newblock \emph{arXiv preprint arXiv:2410.10803}, 2024.

\bibitem[Fang et~al.(2020)Fang, Wang, Gou, and Lu]{fang2020graspnet}
H.-S. Fang, C.~Wang, M.~Gou, and C.~Lu.
\newblock Graspnet-1billion: A large-scale benchmark for general object
  grasping.
\newblock In \emph{Proceedings of the IEEE/CVF conference on computer vision
  and pattern recognition}, pages 11444--11453, 2020.

\bibitem[Fang et~al.(2023)Fang, Wang, Fang, Gou, Liu, Yan, Liu, Xie, and
  Lu]{fang2023anygrasp}
H.-S. Fang, C.~Wang, H.~Fang, M.~Gou, J.~Liu, H.~Yan, W.~Liu, Y.~Xie, and
  C.~Lu.
\newblock Anygrasp: Robust and efficient grasp perception in spatial and
  temporal domains.
\newblock \emph{IEEE Transactions on Robotics}, 2023.

\bibitem[Li et~al.(2023)Li, Liu, Li, Geng, Zhu, Yang, and
  Huang]{li2023gendexgrasp}
P.~Li, T.~Liu, Y.~Li, Y.~Geng, Y.~Zhu, Y.~Yang, and S.~Huang.
\newblock Gendexgrasp: Generalizable dexterous grasping.
\newblock In \emph{2023 IEEE International Conference on Robotics and
  Automation (ICRA)}, pages 8068--8074. IEEE, 2023.

\bibitem[Wang et~al.(2023)Wang, Zhang, Chen, Xu, Li, Liu, and
  Wang]{wang2023dexgraspnet}
R.~Wang, J.~Zhang, J.~Chen, Y.~Xu, P.~Li, T.~Liu, and H.~Wang.
\newblock Dexgraspnet: A large-scale robotic dexterous grasp dataset for
  general objects based on simulation.
\newblock In \emph{2023 IEEE International Conference on Robotics and
  Automation (ICRA)}, pages 11359--11366. IEEE, 2023.

\bibitem[Wen et~al.(2024)Wen, Yang, Kautz, and
  Birchfield]{wen2024foundationpose}
B.~Wen, W.~Yang, J.~Kautz, and S.~Birchfield.
\newblock Foundationpose: Unified 6d pose estimation and tracking of novel
  objects.
\newblock In \emph{Proceedings of the IEEE/CVF Conference on Computer Vision
  and Pattern Recognition}, pages 17868--17879, 2024.

\bibitem[Sundaralingam et~al.(2023)Sundaralingam, Hari, Fishman, Garrett,
  Van~Wyk, Blukis, Millane, Oleynikova, Handa, Ramos,
  et~al.]{sundaralingam2023curobo}
B.~Sundaralingam, S.~K.~S. Hari, A.~Fishman, C.~Garrett, K.~Van~Wyk, V.~Blukis,
  A.~Millane, H.~Oleynikova, A.~Handa, F.~Ramos, et~al.
\newblock Curobo: Parallelized collision-free minimum-jerk robot motion
  generation.
\newblock \emph{arXiv preprint arXiv:2310.17274}, 2023.

\bibitem[Fang et~al.(2019)Fang, Jia, Guo, Xu, Wen, and Sun]{fang2019survey}
B.~Fang, S.~Jia, D.~Guo, M.~Xu, S.~Wen, and F.~Sun.
\newblock Survey of imitation learning for robotic manipulation.
\newblock \emph{International Journal of Intelligent Robotics and
  Applications}, 3:\penalty0 362--369, 2019.

\bibitem[Arunachalam et~al.(2023{\natexlab{a}})Arunachalam, Silwal, Evans, and
  Pinto]{arunachalam2023dexterous}
S.~P. Arunachalam, S.~Silwal, B.~Evans, and L.~Pinto.
\newblock Dexterous imitation made easy: A learning-based framework for
  efficient dexterous manipulation.
\newblock In \emph{2023 ieee international conference on robotics and
  automation (icra)}, pages 5954--5961. IEEE, 2023{\natexlab{a}}.

\bibitem[Arunachalam et~al.(2023{\natexlab{b}})Arunachalam, G{\"u}zey,
  Chintala, and Pinto]{arunachalam2023holo}
S.~P. Arunachalam, I.~G{\"u}zey, S.~Chintala, and L.~Pinto.
\newblock Holo-dex: Teaching dexterity with immersive mixed reality.
\newblock In \emph{2023 IEEE International Conference on Robotics and
  Automation (ICRA)}, pages 5962--5969. IEEE, 2023{\natexlab{b}}.

\bibitem[Zhu et~al.(2023)Zhu, Jiang, Stone, and Zhu]{zhu2023learning}
Y.~Zhu, Z.~Jiang, P.~Stone, and Y.~Zhu.
\newblock Learning generalizable manipulation policies with object-centric 3d
  representations.
\newblock \emph{arXiv preprint arXiv:2310.14386}, 2023.

\bibitem[Zhu et~al.(2024)Zhu, Lim, Stone, and Zhu]{zhu2024vision}
Y.~Zhu, A.~Lim, P.~Stone, and Y.~Zhu.
\newblock Vision-based manipulation from single human video with open-world
  object graphs.
\newblock \emph{arXiv preprint arXiv:2405.20321}, 2024.

\bibitem[Valassakis et~al.(2022)Valassakis, Papagiannis, Di~Palo, and
  Johns]{valassakis2022demonstrate}
E.~Valassakis, G.~Papagiannis, N.~Di~Palo, and E.~Johns.
\newblock Demonstrate once, imitate immediately (dome): Learning visual
  servoing for one-shot imitation learning.
\newblock In \emph{2022 IEEE/RSJ International Conference on Intelligent Robots
  and Systems (IROS)}, pages 8614--8621. IEEE, 2022.

\bibitem[Huang et~al.(2024)Huang, Wang, Li, Zhang, and Fei-Fei]{huang2024rekep}
W.~Huang, C.~Wang, Y.~Li, R.~Zhang, and L.~Fei-Fei.
\newblock Rekep: Spatio-temporal reasoning of relational keypoint constraints
  for robotic manipulation.
\newblock \emph{arXiv preprint arXiv:2409.01652}, 2024.

\bibitem[Kerr et~al.(2024)Kerr, Kim, Wu, Yi, Wang, Goldberg, and
  Kanazawa]{kerr2024robot}
J.~Kerr, C.~M. Kim, M.~Wu, B.~Yi, Q.~Wang, K.~Goldberg, and A.~Kanazawa.
\newblock Robot see robot do: Imitating articulated object manipulation with
  monocular 4d reconstruction.
\newblock \emph{arXiv preprint arXiv:2409.18121}, 2024.

\bibitem[Pan et~al.(2025)Pan, Zhang, Wu, Zhao, Gao, and Dong]{pan2025omnimanip}
M.~Pan, J.~Zhang, T.~Wu, Y.~Zhao, W.~Gao, and H.~Dong.
\newblock Omnimanip: Towards general robotic manipulation via object-centric
  interaction primitives as spatial constraints.
\newblock \emph{arXiv preprint arXiv:2501.03841}, 2025.

\bibitem[Zhou et~al.(2024)Zhou, Su, Chi, Zhang, Wang, Huang, Sheng, and
  Wang]{zhou2024code}
E.~Zhou, Q.~Su, C.~Chi, Z.~Zhang, Z.~Wang, T.~Huang, L.~Sheng, and H.~Wang.
\newblock Code-as-monitor: Constraint-aware visual programming for reactive and
  proactive robotic failure detection.
\newblock \emph{arXiv preprint arXiv:2412.04455}, 2024.

\bibitem[Pari et~al.(2021)Pari, Shafiullah, Arunachalam, and
  Pinto]{pari2021surprising}
J.~Pari, N.~M. Shafiullah, S.~P. Arunachalam, and L.~Pinto.
\newblock The surprising effectiveness of representation learning for visual
  imitation.
\newblock \emph{arXiv preprint arXiv:2112.01511}, 2021.

\bibitem[Lin et~al.(2024)Lin, Hu, Sheng, Wen, You, and Gao]{lin2024data}
F.~Lin, Y.~Hu, P.~Sheng, C.~Wen, J.~You, and Y.~Gao.
\newblock Data scaling laws in imitation learning for robotic manipulation.
\newblock \emph{arXiv preprint arXiv:2410.18647}, 2024.

\bibitem[Liu et~al.(2023)Liu, Zeng, Ren, Li, Zhang, Yang, Li, Yang, Su, Zhu,
  et~al.]{liu2023grounding}
S.~Liu, Z.~Zeng, T.~Ren, F.~Li, H.~Zhang, J.~Yang, C.~Li, J.~Yang, H.~Su,
  J.~Zhu, et~al.
\newblock Grounding dino: Marrying dino with grounded pre-training for open-set
  object detection.
\newblock \emph{arXiv preprint arXiv:2303.05499}, 2023.

\bibitem[Kirillov et~al.(2023)Kirillov, Mintun, Ravi, Mao, Rolland, Gustafson,
  Xiao, Whitehead, Berg, Lo, et~al.]{kirillov2023segment}
A.~Kirillov, E.~Mintun, N.~Ravi, H.~Mao, C.~Rolland, L.~Gustafson, T.~Xiao,
  S.~Whitehead, A.~C. Berg, W.-Y. Lo, et~al.
\newblock Segment anything.
\newblock In \emph{Proceedings of the IEEE/CVF International Conference on
  Computer Vision}, pages 4015--4026, 2023.

\bibitem[De~Haan et~al.(2019)De~Haan, Jayaraman, and Levine]{de2019causal}
P.~De~Haan, D.~Jayaraman, and S.~Levine.
\newblock Causal confusion in imitation learning.
\newblock \emph{Advances in neural information processing systems}, 32, 2019.

\bibitem[Sohn et~al.(2015)Sohn, Lee, and Yan]{sohn2015learning}
K.~Sohn, H.~Lee, and X.~Yan.
\newblock Learning structured output representation using deep conditional
  generative models.
\newblock \emph{Advances in neural information processing systems}, 28, 2015.

\end{thebibliography}


\clearpage
\appendix

\section{Appendix}
\label{sec:appendixA}
\setcounter{figure}{0}
\setcounter{table}{0}
\begin{table}[h]
\renewcommand\arraystretch{1.2}
\centering
\caption{Descriptions of robot dexterous manipulation tasks}
\resizebox{0.95\textwidth}{!}{
\begin{tabular}{cp{11cm}}
\toprule
\textbf{Task} & \multicolumn{1}{c}{\textbf{Task Description}} \\\hline
\textit{Grasp Cup} & The robot needs to use a dexterous hand to grasp the cup on the table. \\\hline
\textit{Take Mug} & The robot needs to use a dexterous hand to take the mug on the table. \\\hline
\textit{Hold Scanner} & The robot needs to use a dexterous hand to hold a barcode scanner from a flat surface, preparing it for use. \\\hline
\textit{Catch Loopy} & The robot needs to use a dexterous hand to catch loopy from the environment. \\\hline
\textit{Pinch Toy} & The robot needs to use a dexterous hand to pinch a small toy, maintaining a gentle but secure grip. \\\hline
\textit{Grasp Sanitizer} & The robot needs to use a dexterous hand to grasp a sanitizer, and hold it in the air. \\\hline
\textit{Lift Tray} & The robot needs to use both dexterous hands to lift a tray, ensuring a stable hold while maintaining balance. \\\bottomrule
\end{tabular}
}\label{task_descriptions}
\end{table}

\section{Appendix}
\label{sec:appendixB}

\begin{small}
\begin{algorithm}[htbp] 
    \caption{Training Process}
    \label{alg:training_algorithm}    
    \KwIn{ Dataset $ \mathcal{D} =\{(\bm{s}_t,\bm{a}_t,)_{i=0}^{n}\}$ }
    \KwOut{ Hand-focus policy $\pi_{hand\_focus}$ }
    \For{each batch}{
        Sample a batch of transitions $(\bm{s}_t,\bm{a}_t)$ from the buffer $\mathcal{D}$\\
        Calculate relative proprioception from $\bm{s}_t$\\
        Calculate relative action chunk from $\bm{a}_t$\\
        Extract hand-focus images from original images in $\bm{s}_t$ \\
        Update the hand-focus policy $\pi_{hand\_focus}$ via Eq.~(\ref{hand_focus_model_training_loss}) \\
    }
\end{algorithm}
\vspace{-0.2cm}
\end{small}

\begin{small}
\begin{algorithm}[htbp] 
    \caption{Inference Process}
    \label{alg:inference_algorithm}    
    \KwIn{ Name of manipulating object, e.g., cup }
    \KwOut{Movement trajectory of dexterous hand}
    \If{ manipulating object exists}{
        locate the manipulating object  via GroundingDINO and SAM \\
        Estimate 6D pose of the manipulating object \\
        Set pre-manipulation pose for robot hand \\
        Generate motion trajectory via CuRobo to arrive at the pre-manipulation pose \\
        Generate manipulation trajectory via the learned object-focus policy \\
    }
\end{algorithm}
\vspace{-0.2cm}
\end{small}


\end{document}